\documentclass{article}
\usepackage{spconf,amsmath,graphicx}
\usepackage[hidelinks]{hyperref}
\usepackage{color}
\usepackage[dvipsnames]{xcolor}
\usepackage[noadjust]{cite}  
\usepackage{cleveref}
\usepackage{multirow}
\usepackage{adjustbox}
\usepackage{siunitx}  
\usepackage{bm}
\usepackage{mathtools}
\usepackage{caption}
\usepackage{setspace}

\graphicspath{{figures/}}

\title{Monotonic segmental attention for automatic speech recognition}
\def\blind{0}

\if\blind1%
\name{BLIND}
\address{BLIND}

\else%

\name{Albert Zeyer$^{1,2}$, Robin Schmitt$^{1,2}$, Wei Zhou$^{1,2}$, Ralf Schlüter$^{1,2}$, Hermann Ney$^{1,2}$}
\address{
  $^1$Human Language Technology and Pattern Recognition,
  Computer Science Department, \\
  RWTH Aachen University, 52062 Aachen, Germany, \\
  $^2$AppTek GmbH, 52062 Aachen, Germany\\
\vspace{-2mm}
\texttt{\scriptsize\{zeyer,zhou,schlueter,ney\}@cs.rwth-aachen.de, robin.schmitt1@rwth-aachen.de}}

\copyrightnotice{978-1-6654-7189-3/22/\$31.00~\copyright2023 IEEE}

\fi%

\newsavebox\CBox
\def\mathBF#1{\sbox\CBox{#1}\resizebox{\wd\CBox}{\ht\CBox}{\ensuremath{\mathbf{#1}}}}
\newcommand{\best}[1]{\mathBF{#1}}  

\newcommand{\argmax}{\operatornamewithlimits{argmax}}

\renewcommand{\paragraph}[1]{\noindent\textbf{#1}\ \ }

\makeatletter

\renewcommand{\section}{\@startsection
  {section}%
  {1}%
  {}%
  {-0.5\baselineskip}%
  {0.3\baselineskip}%
  {}}%

\renewcommand{\subsection}{\@startsection
  {subsection}%
  {2}%
  {}%
  {-0.4\baselineskip}%
  {0.2\baselineskip}%
  {}}%

\renewcommand{\subsubsection}{\@startsection
  {subsubsection}%
  {3}%
  {}%
  {-0.1\baselineskip}%
  {0.1\baselineskip}%
  {}}%

\g@addto@macro\normalsize{%
  \setlength\abovedisplayskip{4pt plus 2pt minus 2pt}
  \setlength\belowdisplayskip{4pt plus 2pt minus 2pt}
  \setlength\abovedisplayshortskip{4pt plus 2pt minus 2pt}
  \setlength\belowdisplayshortskip{4pt plus 2pt minus 2pt}
}

\begin{document}

\maketitle

\begin{abstract}
We introduce a novel segmental-attention model
for automatic speech recognition.
We restrict the decoder attention to segments
to avoid quadratic runtime of global attention,
better generalize to long sequences,
and eventually enable streaming.
We directly compare global-attention
and different segmental-attention modeling variants.
We develop and compare
two separate time-synchronous decoders,
one specifically taking the segmental nature into account,
yielding further improvements.
Using time-synchronous decoding for segmental models is novel
and a step towards streaming applications.
Our experiments show the importance of a length model
to predict the segment boundaries.
The final best segmental-attention model using segmental decoding
performs better than global-attention,
in contrast to other monotonic attention approaches in the literature.
Further, we observe that the segmental model
generalizes much better to long sequences of up to several minutes.
\end{abstract}


\section{Introduction \& related work}  

The attention-based encoder-decoder architecture \cite{bahdanau2015nmt}
has been very successful as an end-to-end model
for many tasks including speech recognition
\cite{chorowski2015attention,chan2016las,%
zeyer2018:asr-attention,park2019specaugment,tuske2020swbatt}.
However, for every output label,
the attention weights are over all the input frames,
referred to as \emph{global attention}.
This has the drawbacks of quadratic runtime-complexity,
potential non-monotonicity;
it does not allow for online streaming recognition
and it does not generalize to longer sequence lengths than those seen in training
\cite{rosendahl2019:pos_enc,narayanan2019longform,chiu2019longform}.

There are many attempts to solve some of these issues.
Monotonic chunkwise attention (MoChA)
\cite{chiu2018monotonic,fan2018mocha,arivazhagan2019milk}
is one popular approach which uses fixed-size chunks
for soft attention
and a deterministic approach for the chunk positions,
i.e.~the position is not treated as a latent variable in recognition.
Many similar approaches using a local fixed-sized window
and some heuristic or separate neural network for the position prediction
were proposed
\cite{chorowski2015attention,bahdanau2016att,%
luong2015attentionmt,tjandra2017localmonotonicatt,zhang2019windowedatt,%
jaitly2016onlineseq2seq,raffel2017onlinemonotonic,%
merboldt2019:local-monotonic-att,%
Hsiao2020OnlineLAS}.
The attention sometimes also uses a Gauss distribution
which allows for a differentiable position prediction
\cite{graves2013generatingseqs,hou2017gaussian,%
luong2015attentionmt,%
tjandra2017localmonotonicatt,%
zhang2019windowedatt}.
Some models add a penalty in training,
or are extended to have an implicit bias to encourage monotonicity
\cite{chorowski2014end2end,chorowski2015attention,bahdanau2016att}.
Framewise defined models like CTC \cite{graves2006connectionist}
or transducers \cite{graves2012seqtransduction}
canonically allow for streaming,
and there are approaches to combine such framewise model with attention
\cite{watanabe2017hybridctcatt,hori2017jointctcatt,kim2017jointctcatt,%
moritz2019triggeredatt,moritz2019ctcatt,%
miao2019mocha,chiu2019twopass,sainath2020streaming}.

Our probabilistic formulation using latent variables for the segment boundaries
is similar to other segmental models
\cite{ostendorf1996hmm,%
kong2015segmental,lu2016segmental,%
yu2016onlines2s,yu2016noisychannel,%
DoetschHannemannSchlueter17:InvHMM,beck2018:segmental,beck2018:alignment,%
wang17:directhmm,wwang18:neuralhmm,%
alkhouli16:align,alkhouli2020:phd,%
zeyer2021:latent-attention},
although attention has not been used on the segments except in
\cite{zeyer2021:latent-attention}
and there are usually more independence assumptions
such as first or zero order dependency on the label,
and only a first order dependency on the segment boundary,
which is a critical difference.
It has also been shown that transducer models and segmental models
are equivalent \cite{zhou2021segmentalvstransducer}.

Here, we want to make use of the attention mechanism
while solving the mentioned global attention drawbacks
by making the attention \emph{local} and \emph{monotonic} on \emph{segments}.
We treat the segment boundaries as latent variables
and end up at our segmental attention models.
Such segmental models by definition are monotonic, allow for streaming,
and are much more efficient by using only local attention.
Our aim is to get a better understanding of such segmental attention models,
to directly compare it to the global attention model,
to study
how to treat and model the latent variable,
how to perform the search for recognition,
how to treat silence,
how well it generalizes to longer sequences
among other questions.

\section{Global attention model}

Our starting point is the
standard global-attention-based encoder-decoder model \cite{bahdanau2015nmt}
adapted for speech recognition
\cite{chorowski2015attention,chan2016las,%
park2019specaugment,tuske2020swbatt},
specifically the model as described in
\cite{zeyer2018:asr-attention,zeyer2019:trafo-vs-lstm-asr,zeyer2020:transducer}.
We use an LSTM-based \cite{hochreiter1997lstm} encoder
which gets a sequence of audio feature frames $x_1^{T'}$ of length $T'$ as input
and encodes it as a sequence
\[ h_1^T = \operatorname{Encoder}(x_1^{T'}) \]
of length $T$, where we apply downsampling by max-pooling in time
inside the encoder by factor 6.
For the output label sequence $a_1^S$ of length $S$,
given the encoder output sequence $h_1^T$ of length $T$,
we define
\begin{align*}
p(a_1^S \mid h_1^{T})
&=
\prod_{s=1}^{S}
\underbrace{p(a_s \mid a_1^{s-1}, h_1^T)}_\text{label model} ,
\end{align*}
The label model uses global attention on $h_1^T$ per each output step $s$.
The neural structure which defines the probability distribution of the labels
is also called the decoder.
The decoder of the global-attention model
is almost the same as
our segmental attention model, which we will define below in detail.
The main difference is segmental vs.~global attention.

\section{Our segmental attention model}

\begin{figure}
\includegraphics[width = 0.9\columnwidth]{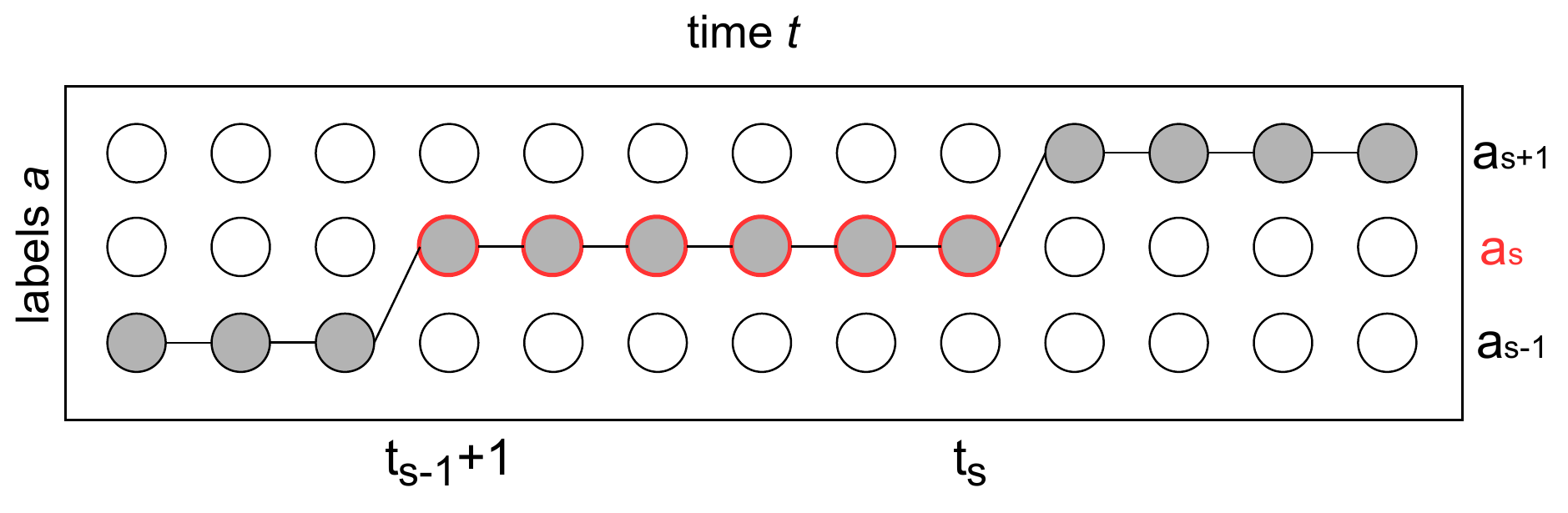}
\caption{Example alignment of label sequence $...a_{s-1}a_sa_{s+1}...$. Segment for $a_s$ highlighted.}
\label{fig:segment-vis}
\end{figure}

\begin{figure}
\centering
\includegraphics[width = 0.55\columnwidth]{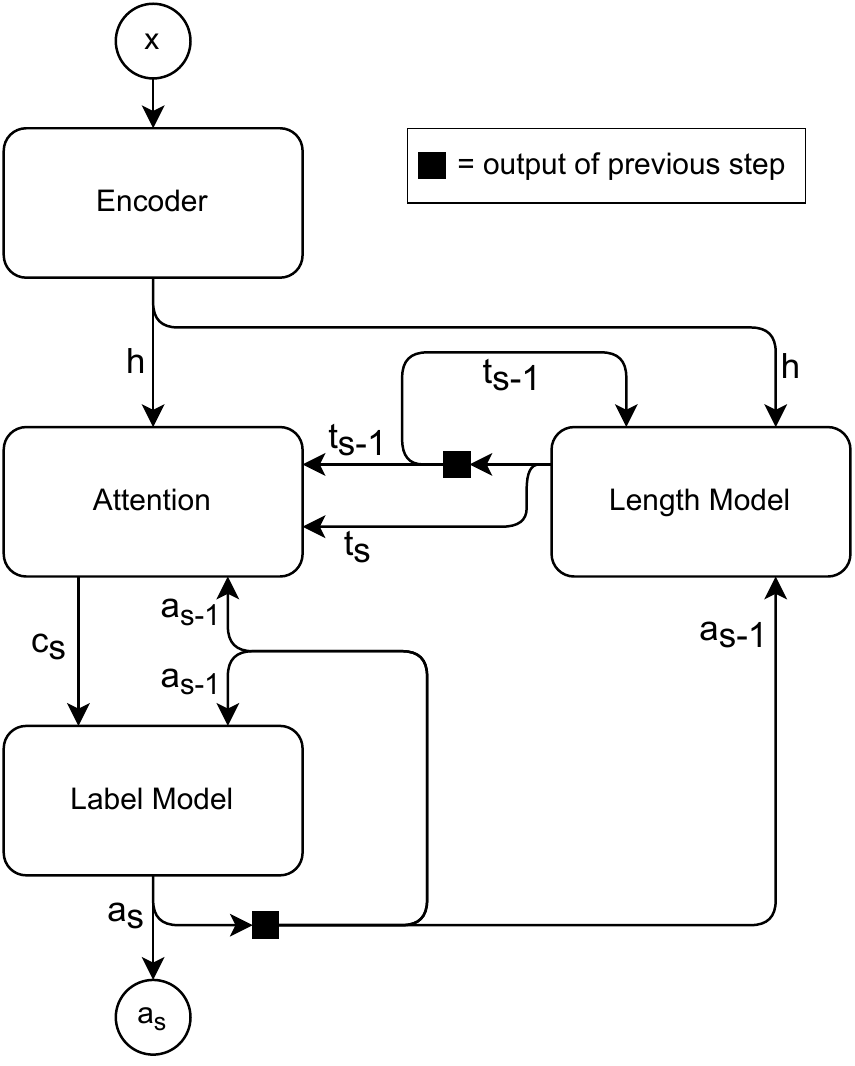}
\caption{Segmental attention model}
\label{fig:segmental-model-overview}
\end{figure}

Now, we introduce \emph{segment boundaries}
as a sequence of \emph{latent variables} $t_1^S$.
Specifically, for one output $a_s$, the segment is defined by $[t_{s-1} + 1, t_s]$,
with $t_0 = 0$ and $t_S = T$ fixed,
and we require $t_s > t_{s-1}$ for \emph{strict monotonicity}.
Thus, the segmentation fully covers all frames of the sequence.
One such segment is highlighted in \Cref{fig:segment-vis}.
The label model now uses attention only in that segment,
i.e.~on $h_{t_{s-1}+1}^{t_s}$.
For the output label sequence $a_1^S$, 
we now define the segmental model as 
\begin{align*}
p(a_1^S \mid h_1^{T})
&=
\sum_{t_1^S} \prod_{s=1}^{S}
\underbrace{p(t_s \mid ... \vphantom{h_{t_{s-1}+1}^{t_s}})^\alpha}_\text{length model} \cdot
\underbrace{p(a_s \mid a_1^{s-1}, h_{t_{s-1}+1}^{t_s}, ...)}_\text{label model}
.
\end{align*}
%
In the simplest case, we even do not use any length model ($\alpha = 0$).
The intuition was that a proper dynamic search over the segment boundaries
can be guided by the label model alone,
as it should produce bad scores for bad segment boundaries.
We also test a simple static length model,
and a neural length model,
as we will describe later.
The label model is mostly the same as in the global attention case
with the main difference that we have the attention only on $h_{t_{s-1}+1}^{t_s}$.
The whole segmental model is depicted in \Cref{fig:segmental-model-overview}.

\subsection{Label model variations}
\label{sec:label-model}

\begin{figure}
\centering
\includegraphics[width = .9\columnwidth]{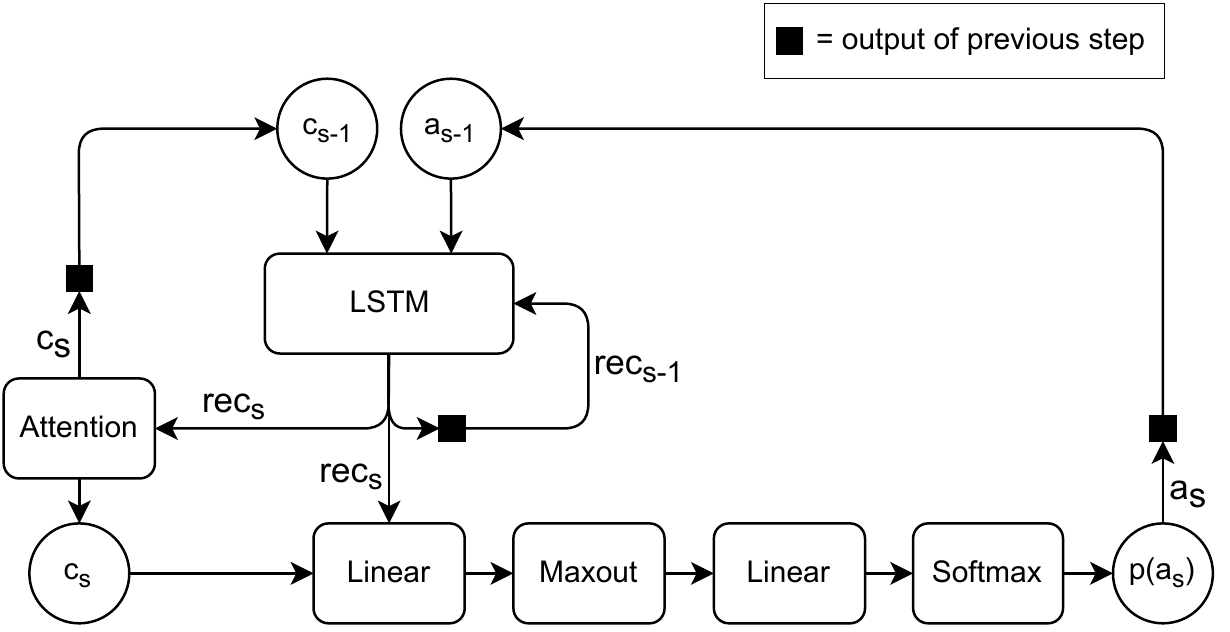}
\caption{Segmental attention label model $p(a_s | ...)$}
\label{fig:segmental-label-model}
\end{figure}

\newcommand{\statehighlight}[1]{{\color{blue}\bm{#1}}}
\newcommand{\atthighlight}[1]{{\color{RubineRed}\bm{#1}}}

The label model is depicted in \Cref{fig:segmental-label-model}
and defined as
\begin{align*}
p(a_s \mid ...)
&= (\operatorname{softmax} \circ \operatorname{Linear} \circ \operatorname{maxout} \circ \operatorname{Linear}) \big( \\
&\quad \quad (\operatorname{LSTM}(\statehighlight{c_1^{s-1}}, a_1^{s-1}),
c_s
) \big)
\\
c_s
&= \sum_{t=\atthighlight{t_{s-1}+1}}^{\atthighlight{t_s}}
\alpha_{s,t}
\cdot h_t \\
\alpha_{s,t}
&=
\frac{\exp (e_{s,t})}{\sum_{\tau=\atthighlight{t_{s-1}+1}}^{\atthighlight{t_s}} \exp (e_{s,\tau})} \\
e_{s,t}
&=
(\operatorname{Linear} \circ \tanh \circ \operatorname{Linear}) \big( \\
&\quad\quad (\operatorname{LSTM}(\statehighlight{c_1^{s-1}}, a_1^{s-1}), h_t) \big) .
\end{align*}
The attention weights here are only calculated on the interval $[\atthighlight{t_{s-1}+1},\atthighlight{t_s}]$.
Further, we do not have attention weight feedback as there is no overlap between the segments.
Otherwise the model is exactly the same as the global attention decoder,
to allow for direct comparisons, and also to import model parameters.

Another variation is on the dependencies.
In any case, we depend on the full label history $a_1^{s-1}$.
When we have the dependency on $\statehighlight{c_1^{s-1}}$ as it is standard
for LSTM-based attention models,
this implies an implicit dependency on the whole past segment boundaries $t_1^{s-1}$,
which removes the option of an exact first-order dynamic programming implementation
for forced alignment or exact sequence-likelihood training.
So, we also test the variant where we remove the $\statehighlight{c_1^{s-1}}$ dependency
in the equations above.

\subsection{Silence modeling}

The output label vocabulary of the global attention model usually does not include silence
as this is not necessarily needed.
Our segments completely cover the input sequence,
and thus the question arises whether the silence parts should be separate segments or not,
i.e.~whether we should add silence to the vocabulary.
Additionally, as silence segments tend to be longer,
we optionally split them up.
We will perform experiments for all three variants, also for the global attention model.

Further, when we include silence,
the question is whether this should be treated just as another output label,
or treated in a special way, e.g.~separating it from the $\operatorname{softmax}$.
We test some preliminary variations on this.

\subsection{Length model variations and length normalization}

\paragraph{No length model.}
The simplest variant is without any length model ($\alpha=0$).
We might argue that this is closest to the global attention model.
In this case, we use length normalization instead during recognition
as explained below,
just like we do for the global attention model.

\vspace{0.4\baselineskip}
\paragraph{Static length model.}
Here we estimate a model
\begin{align*}
\omit\rlap{$p(t_s = t_{s-1} + \Delta t \mid a_s, t_{s-1})$} & \\
&= p(\Delta t \mid a_s)
=
\begin{cases}
  \exp(-|\mu_{a_s} - \Delta t|)/Z, & 1 \le \Delta t \le \delta_\text{max} \\
  0, & \text{otherwise}
\end{cases}
\end{align*}
with mean segment length $\mu$,
\newcommand{\maxseglen}{\delta_\text{max}}
maximum segment length $\maxseglen$ and normalization $Z$,
where $\mu$ is estimated
based on some given alignments.
Note that this invalidates a proper probability normalization
when combined with the label model $p(a_s | ...)$.
However, we can interpret this as a log-linear combination
of the joint model $p(t_s,a_s | ...)$,
although we do not use a proper renormalization here.

\vspace{0.4\baselineskip}
\paragraph{Neural length model.}
\label{sec:neural-length-model}
This model is defined as
\newcommand{\alignlabel}{\omega}
\newcommand{\blank}{\left<\operatorname{b}\right>}
\begin{align*}
%
\omit\rlap{$p(t_s = t \mid a_1^{s-1}, t_1^{s-1}, h_1^T)$} &
\\& =
\begin{cases}
0,  &  t < t_{s-1}+1 \\
%
\left[
\prod_{t'=t_{s-1}+1}^{t-1}
1 - q_{t'}
\right]
\cdot q_{t},
& t \ge t_{s-1}+1
\end{cases}
\\
q_{t'} &=
(\operatorname{sigmoid} \circ \operatorname{Linear} \circ \tanh \circ \operatorname{LSTM})
(\alignlabel_0^{t' - 1}, h_1^{t'})
\end{align*}
The $q_{t'}$ model here is defined in a framewise manner
and predicts whether the new segment ends in the frame $t'$,
similar as in
\cite{yu2016onlines2s,yu2016noisychannel,%
beck2018:segmental,%
chiu2018monotonic,%
zeyer2020:transducer,zhou2021segmentalvstransducer}.
This $q_{t'}$ model gets in the framewise alignment $\omega_0^{t'-1}$ so far,
where we encode the labels $a_1^{s-1}$
at the segment boundaries $t_1^{s-1}$
and encode a special blank symbol $\blank$ otherwise,
specifically
\begin{align*}
\alignlabel_{t''} & =
\begin{cases}
a_{s'}, & t_{s'} = t'' \\
\blank, & \text{otherwise}
\end{cases}
.
\end{align*}

\subsection{Training}

For simplicity,
we use framewise cross entropy (CE) as training criterion
using a given framewise alignment.
Other from-scratch training schemes are possible
without the requirement of a given alignment
but we do not investigate this further here.
We use alignments from a HMM
for the variants with silence
and alignments from a RNA model \cite{sak2017neuralaligner} otherwise.
This defines the segment boundaries $t_s$.
We minimize the loss
\[
  L(\theta) = \sum_{s=1}^S
    - \log p_\theta(t_s \mid ...)
    - \log p_\theta(a_s \mid ...)
\]
w.r.t.~the model parameters $\theta$ over the training data.

For the segmental-attention model, we can also directly use the model parameters
of a global-attention model, as it is just the same model
except that the attention is limited to the segment,
and except of the length model.
We also perform such experiments.

\subsection{Recognition: Time-sync.~decoding \& pruning}
\label{sec:recog}

For the segmental model,
our decision rule is
\begin{align*}
x_1^{T'}
\mapsto
&
\ \hat{S},\hat{a}_1^{\hat{S}},\hat{t}_1^{\hat{S}}
:=
\argmax_{S,a_1^S,t_1^S}
\left( \vphantom{\sum_{s=1}^S} \right. \\
&
\left.
\sum_{s=1}^S
    \alpha \cdot \log p_\theta(t_s \mid ...)
    + \log p_\theta(a_s \mid ...)
\right)
\cdot S^{-\gamma}
.
\end{align*}
We test different length model scales $\alpha$
and we either enable or disable length normalization
($\gamma = 0,1$)
\cite{murray2018lengthbias}.
We don't use an external language model in this work.

We use time synchronous decoding,
meaning that the
state extension and pruning
is synchronized in a frame-by-frame manner,
i.e.~all active hypotheses are in the same time frame.
We have two separate decoder implementations for our segmental model,
both of them being time-synchronous but differing in the pruning variant.
They are both based on our time-synchronous transducer decoding
and use a reformulation of the segmental model as a transducer model
similar to \cite{zhou2021segmentalvstransducer}.
The label model contribution is added at the end of a segment.
In case of the neural length model,
we can use $q$ for a framewise contribution,
or otherwise add the length model contribution at the end of a segment.

\vspace{0.4\baselineskip}
\paragraph{Segmental search.}
At each time frame,
we only prune those hypotheses reaching segment end
against each other,
while others are kept for further extension
unless they reach a maximum segment length $\maxseglen$.
We use Viterbi recombination
of hypotheses with the same label history.
This implementation is more complex and runs on CPU.

\vspace{0.4\baselineskip}
\paragraph{Simple search.}
At each time frame,
we prune all hypotheses together
regardless of reaching segment end or not.
This is done only with the neural length model,
where ongoing segments still have an intermediate length probability (framewise $q_t$)
for pruning.
This is exactly the transducer search without any recombination.
The algorithm is simple and
we have an efficient GPU-based implementation.

\vspace{0.4\baselineskip}
This is in contrast to label-synchronous search as it is done usually for
global attention models or also other segmental models
\cite{zeyer2021:latent-attention},
where all current active hypotheses have the same number of output labels.

\section{Experiments \& analysis}

We perform all our experiments on the
Switchboard 300h English telephone speech corpus \cite{godfrey1992switchboard}.
We use a vocabulary of 1030 byte pair encoding (BPE) \cite{sennrich16bpe} tokens
with optional silence.
The encoder is always a 6-layer bidirectional LSTM encoder of 1024 dimensions in each direction
with downsampling factor 6 via max-pooling and has 137M parameters.
The decoder uses a single LSTM of 1024 dimensions
and has 24M parameters without the length model
and is mostly identical for global-attention and segmental-attention.
The length model has 1.3M parameters.
\if\blind1%
We implement the neural models and the search for recognition
in our internal toolkit.
We will release all the code and recipes after acceptance.
\else%
We implement the neural models and the simple search for recognition
in RETURNN \cite{zeyer2018:returnn}
and the segmental search in RASR \cite{rybach2011:rasr}.
All the code and recipes can be found online%
\footnote{\tiny\url{https://github.com/rwth-i6/returnn-experiments/tree/master/2022-segmental-att-models}}.
\fi%


We also provide the number of search errors by counting the number of sequences
where the ground truth sequence has a higher score than the recognized sequence,
as a sanity check to indicate how much errors we make due to the approximations in the search.

\subsection{Failures when no length model is used}


\begin{table}[t]
\caption{\textbf{Global vs.~segmental attention} without length model but length normalization.
Word-error-rate (WER) on Hub5'00 with substitutions (S), deletions (D) and insertions (I),
and search errors (SE).
All models with silence; segmental search for segmental models.}
\label{tab:global-vs-seg-initial}
\centering
\setlength{\tabcolsep}{0.3em}
\begin{tabular}{|c|c|S[table-format=1.1]|S[table-format=2.1]|S[table-format=1.1]||S[table-format=2.1]|S[table-format=1.1]|}
  \hline
  Model & Parameters & \multicolumn{4}{c|}{WER (\%)} & {\multirow{2}{*}{\shortstack{SE\\(\%)}}} \\
  & & {S} & {D} & {I} & {$\Sigma$} & \\
  \hline
  \hline
  Global & Trained & 9.3 & 4.6 & 2.0 & \best{15.9} & 1.0 \\ \hline
  \multirow{2}{*}{Segmental} & From global & 6.7 & 35.8 & 6.3 & 48.8 & 1.0 \\ \cline{2-7}
  & Trained & 6.4 & 17.2 & 1.7 & 25.3 & 0.4 \\
  \hline
\end{tabular}
\end{table}

Some initial results comparing global and segmental attention
are shown in \Cref{tab:global-vs-seg-initial}.
This is without a length model but with length normalization.
For better comparison, all models include silence labels.
The segmental models perform badly.
In the following, we analyze why that is.

\begin{table}
\caption{The \textbf{model scores} $\log p(a_s | ...)$ (higher is better)
for the global and segmental attention models
over the ground truth and recognized sequence per label and in total
for an example sequence.}
\label{tab:global-restricted-deletions-label-scores}
\begin{adjustbox}{width=0.9\width,center}
\begin{tabular}{|l|S[table-format=1.2]|S[table-format=1.2]|S[table-format=1.2]|}
    \hline
    \multirow{2}{*}{Labels} & {Global attention} & \multicolumn{2}{c|}{Segmental attention} \\
    \cline{2-4}
    & {\shortstack{Ground-truth\\(= recognized)}} & {Ground-truth} & {Recognized} \\
    \hline    \hline
    [Silence] & -0.11 & -0.11 & -0.13 \\    \hline
    [Silence] & -0.10 & -0.10 & -0.07 \\    \hline
    [Silence] & -0.11 & -0.11 & -0.09 \\    \hline
    you & -0.09 & -0.06 & {\multirow{2}{*}{(Deletion)}} \\    \cline{1-3}
    know & -0.25 & -0.10 & \\    \hline
    gra@@ & -0.06 & -0.31 & -0.01 \\    \hline
    de & -0.13 & -0.02 & -0.03 \\    \hline
    school & -0.02 & -0.06 & -0.04 \\    \hline
    or & -0.23 & -0.01 & -0.04 \\    \hline
    so & -0.10 & -0.12 & -0.08 \\    \hline
    \multicolumn{4}{c}{...} \\    \hline
    $\Sigma$ & -1.71 & -1.44 & -0.54\\    \hline
    $\Sigma / S$ & -0.11 & -0.10 & -0.06 \\    \hline
\end{tabular}
\end{adjustbox}
\end{table}

As a first step, we want to understand the behavior
when we take over the parameters of a global attention model
and just restrict the attention on given segment boundaries.
We now look at individual scores per label.
We chose a sequence where the global attention model
recognizes exactly the ground truth label sequences
but the segmental model recognizes some different sequence.
The scores are in \Cref{tab:global-restricted-deletions-label-scores}.
We see that restricting the attention to the segment
actually improves the recognition score slightly ($-0.10 > -0.11$).
However, now the segmental models recognizes another bad sequence
with even better scores ($-0.06$).
So the score of the correct sequence improved
but the score of incorrect sequences improved even more.

\begin{figure}[t]
\centering
\includegraphics[width=\columnwidth]{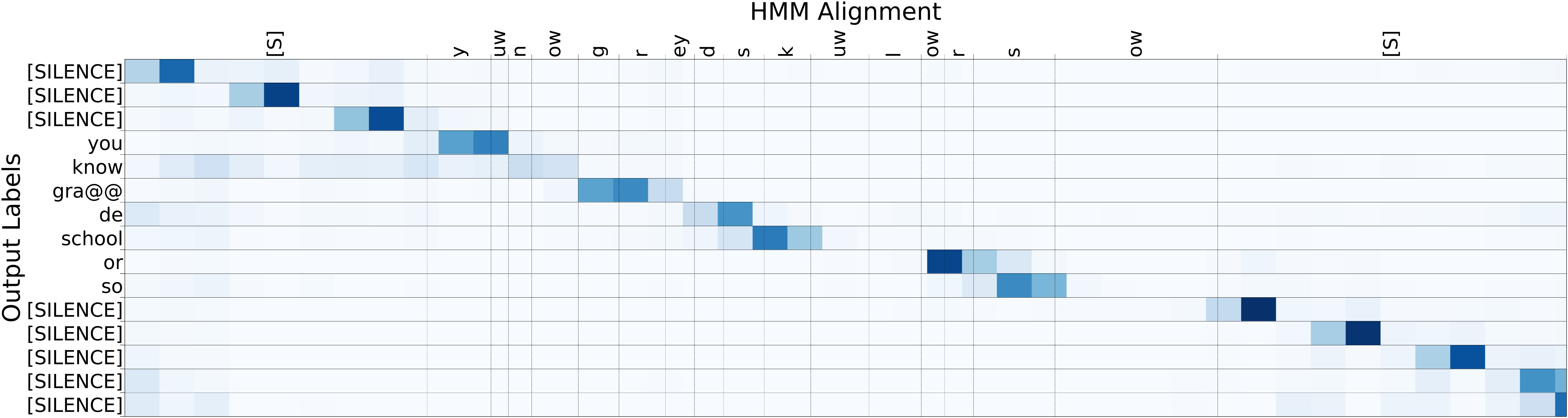}
\includegraphics[width=\columnwidth]{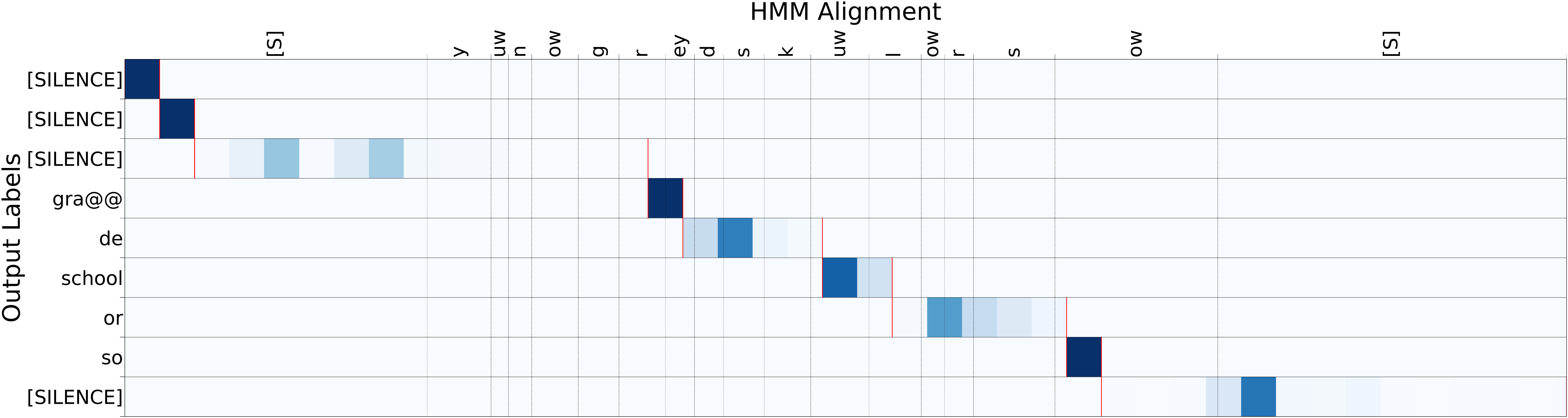}
\caption{Top: Attention weights of the global attention model.
Bottom: Segment boundaries and attention weights of the segmental model.
We also show the reference HMM alignment.}
\label{fig:seg-boundaries-no-length-model}
\end{figure}

We also looked at the segment boundaries of the recognized sequence.
The segment boundaries together with the attention weights
can be seen in \Cref{fig:seg-boundaries-no-length-model}.
We notice that the segment boundaries are very off
and they often cover multiple labels
which explains the high deletion errors from \Cref{tab:global-vs-seg-initial}.
This observation is consistent over many other examples as well.
We also make the same observations when we train the segmental model
in contrast to importing the global attention model parameters.

To summarize: the label model yields good scores even for bad segment boundaries.
This is probably due to the attention mechanism.
This demonstrates that we need some sort of length model
to get better segment boundaries.

\subsection{Length model comparison}


\begin{table}
\caption{Comparing different \textbf{length models}, with optional length normalization.
Segmental search; models with silence.}
\centering
\label{tab:segmental-variants-length-model-compare}
\setlength{\tabcolsep}{0.4em}
\begin{tabular}{|c|c|S[table-format=2.1]|S[table-format=2.1]|S[table-format=1.1]||S[table-format=2.1]|S[table-format=1.1]|}
  \hline
  \multirow{2}{*}{\shortstack{Length\\model}} &
  \multirow{2}{*}{\shortstack{Length\\norm.\vphantom{l}}} &
  \multicolumn{4}{c|}{WER (\%)} &
  {\multirow{2}{*}{\shortstack{SE\\(\%)}}} \\
  && {S} & {D} & {I} & {$\Sigma$} & \\
  \hline
  \hline
  \multirow{2}{*}{None} & No & 5.3 & 29.7 & 0.4 & 35.4 & 0.0 \\ \cline{2-7}
  & Yes & 6.4 & 17.2 & 1.7 & 25.3 & 0.4 \\ \hline
  \multirow{2}{*}{Static} & No & 11.9 & 18.9 & 3.8 & 34.5 & 0.0 \\ \cline{2-7}
   & Yes & 10.2 & 16.3 & 5.1 & 31.7 & 0.0 \\ \hline
  \multirow{2}{*}{Neural} & No & 9.0 & 6.2 & 1.9 & \best{17.1} & 0.0 \\ \cline{2-7}
  & Yes & 9.1 & 5.9 & 2.0 & \best{17.0} & 0.0 \\ \hline
\end{tabular}
\end{table}

In all following experiments,
we always train the segmental model and do not import weights.

We compare our different length models and length normalization
in \Cref{tab:segmental-variants-length-model-compare}.
As expected, without any length model,
the length normalization improves the performance.
For the other length models, we use scale $\alpha=1$.
We notice that the simple static length model yields even worse results
than no length model.
We analyze that in the following subsection and a tuned $\alpha$ will improve that.
Overall, the neural length model yields the best results,
where the length normalization has a negligible effect.
In better segmental models with neural length model
as presented later (e.g.~\Cref{tab:final-results}),
length normalization was hurtful and we did not use it.
To conclude, length normalization is only needed without or with a weak length model.

\subsection{Failures with static length model}

\begin{table}[t]
\caption{\textbf{Segmental model scores} using static length model:
Label scores $\log p(a_s | ...)$ and static length model $\log p(t_s | ...)$.}
\centering
\label{tab:segmental-static-length-model-label-scores}
\begin{adjustbox}{width=0.9\width,center}
\setlength{\tabcolsep}{0.3em}
\begin{tabular}{|l|S[table-format=1.2]|S[table-format=1.2]|S[table-format=2.2]|S[table-format=2.2]|}
  \hline
  \multirow{2}{*}{Labels} & \multicolumn{2}{c|}{Label model scores} & \multicolumn{2}{c|}{Length model scores} \\
  \cline{2-5}
  & {Ground-truth} & {Recognized} & {Ground-truth} & {Recognized} \\
  \hline  \hline
  [Silence] & -0.11 & -0.11 & -1.00 & -1.00 \\  \hline
  [Silence] & -0.10 & -0.10 & -1.00  & -1.00  \\  \hline
  parti@@ & -0.10 & -0.08 & -1.47 & -0.84 \\  \hline
  cu@@ & -0.05 & -0.04 & -0.88 & -0.88 \\  \hline
  lar@@ & -0.06 & -1.18 & -1.57 & -0.79 \\  \hline
  ly & -0.05 & -0.14 & -1.16 & -1.02 \\  \hline
  if & -0.02 & {\multirow{2}{*}{(Deletion)}} & -1.59 & {\multirow{2}{*}{(Deletion)}} \\  \cline{1-2}\cline{4-4}
  you're & -0.61 &  & -1.60 &  \\  \hline
  h@@ & -0.27 & -0.54 & -0.81 & -1.40 \\  \hline
  un@@ & -0.06 & -0.31 & -0.89 & -0.89 \\  \hline
  g@@ & -0.01 & -0.20 & -1.67 & -0.67 \\  \hline
  ry & -0.03 & -0.05 & -1.03 & -1.03 \\  \hline
  \multicolumn{5}{c}{...} \\  \hline
  $\Sigma$ & -1.75 & -2.90 & -17.85 & -10.52 \\
  \hline
\end{tabular}
\end{adjustbox}
\end{table}

We want to better understand why the static length model fails
and whether it fails for similar reasons as without any length model.
We again look into the model scores for an example sequence
in \Cref{tab:segmental-static-length-model-label-scores}.
Now we notice a new problem:
The length model scores dominate much over the label scores.
The label model scores of the ground-truth sequence
is actually better than for the recognized sequence.
However, the length model dominates and prefers some incorrect
segment boundaries.

\begin{table}
\caption{Static length model \textbf{scales} ($\alpha$)}
\centering
\label{tab:static-length-model-scales}
\begin{tabular}{|c||c|c|c|c|c|}
\hline
Length model scale & $0.0$  & $0.01$ & $0.1$  & $0.3$  & $1.0$  \\ \hline
WER (\%)           & $25.3$ & $23.9$ & $\best{21.8}$ & $24.9$ & $31.7$ \\ \hline
\end{tabular}
\end{table}

To overcome this, we test different length model scales $\alpha$
in \Cref{tab:static-length-model-scales},
to transition from no length model to the static length model.
We see that we can still improve over the cases $\alpha = 0$ or $\alpha = 1$
but we are still behind the global attention model.
To summarize: A weak length model also leads to suboptimal results.
Thus we will focus now on the segmental model with neural length model.

\subsection{Silence variants}

\begin{table}
\caption{Comparing \textbf{silence variants}.}
\centering
\label{tab:silence-variants}
\begin{adjustbox}{width=1.\width,center}
\setlength{\tabcolsep}{0.5em}
\begin{tabular}{|c|c|S[table-format=1.1]|S[table-format=1.1]|S[table-format=1.1]||S[table-format=2.1]|S[table-format=1.1]|}
  \hline
  Model & \multirow{2}{*}{\shortstack{Silence\\variant}} &
  \multicolumn{4}{c|}{WER (\%)} & {\multirow{2}{*}{\shortstack{SE\\(\%)}}} \\
  & & {S} & {D} & {I} & {$\Sigma$} & \\
  \hline
  \hline
  \multirow{3}{*}{Global} & None & 9.2 & 3.3 & 2.2 & 14.8 & 0.7 \\
  \cline{2-7}
  & No split & 9.5 & 3.4 & 2.1 & 15.0 & 0.6 \\
  \cline{2-7}
  & Split & 9.3 & 4.6 & 2.0 & 15.9 & 1.0 \\
  \hline
  \multirow{2}{*}{Segmental} & None & 8.7 & 3.5 & 1.9 & \best{14.1} & 0.0 \\
  \cline{2-7}
  & Split & 9.0 & 6.2 & 1.9 & 17.1 & 0.0 \\
  \hline
\end{tabular}
\end{adjustbox}
\end{table}

We compare the standard global-attention case
of not having silence
vs.~having silence,
either being split up into multiple segments in case they are too long,
or not split up.
The information on silence is extracted from our HMM alignment.
The results are in \Cref{tab:silence-variants}.
The variant without silence is better in all cases.
Without silence,
now the segmental attention model performs even better than the global attention model.
%
We think that in principle it should be possible
to improve the variant with silence
and we have some ongoing work on this.



\subsection{Search comparison}

\begin{table}
\caption{\textbf{Search comparison} of simple search vs.~segmental search,
using segmental models with silence or without.
{\footnotesize *Not directly comparable because of CPU vs.~GPU.}}
\centering
\label{tab:search-compare}
\begin{tabular}{|c|S[table-format=2.1]|S[table-format=1.1]|S[table-format=1.1]|S[table-format=2.1]|S[table-format=1.1]|S[table-format=2.1]|}
    \hline
    Variant & \multicolumn{3}{c|}{Simple search} & \multicolumn{3}{c|}{Segmental search}\\
    \cline{2-7}
    & {WER} & {SE} & {Time} & {WER} & {SE} & {Time} \\
    & {(\%)} & {(\%)} & {(h)} & {(\%)} & {(\%)} & {(h)} \\
    \hline
    \hline
    W/o~sil. & 15.0 & 0.8 & 0.1* & 14.1 & 0.0 & 19.5* \\
    \hline
    W/ sil. & 16.4 & 2.4 & 0.1* & 17.1 & 0.0 & 11.6*  \\
    \hline
\end{tabular}
\end{table}

Results of our two search implementations
as described in \Cref{sec:recog}
are collected in \Cref{tab:search-compare},
comparing the same segmental model.
Without silence, we get better performance with the segmental search.
With silence, the result is unexpected.
We can see that segmental search even has less search errors,
so it looks like the model is worse
but this needs further investigation,
as also already mentioned in the last subsection.
Segmental search is slower than simple search,
although the numbers are not directly comparable
due to CPU vs.~GPU.
Compared to the global attention model,
for the segmental search,
the computational complexity increases by a constant factor
by the maximum segment length $\maxseglen$,
but the attention computation complexity decreases from $T \cdot S$ to $\maxseglen \cdot S$,
which dominates once $T$ becomes big.

\subsection{State vector type}

\begin{table}
\caption{Testing \textbf{state vector dependency} on $c_1^{s-1}$.
Segmental model uses segmental search, neural length model.}
\label{tab:state-vector}
\centering
\begin{tabular}{|c|c|c|c|}
  \hline
  Model & $c_1^{s-1}$ dependency & WER (\%) & SE (\%) \\
  \hline
  \hline
  \multirow{2}{*}{Global} & Yes & 15.3 & 0.5 \\
  \cline{2-4}
  & No & 16.8 & 0.5 \\
  \hline
  \multirow{2}{*}{Segmental} & Yes & 14.5 & 0.0 \\
  \cline{2-4}
  & No & 14.9 & 0.0 \\
  \hline
\end{tabular}
\end{table}

As described in \Cref{sec:label-model},
we have two variants of the state vector,
either with dependency on $c_1^{s-1}$ or without.
Results are in \Cref{tab:state-vector}.
As expected, the global attention model
benefits a lot by having this dependency,
as it can use it to better guide where to attend next.
The difference is much smaller for the segmental model,
as the attention is anyway only on the segment.
However, it still slightly benefits from having the dependency.

\subsection{Generalization on longer sequences}

\begin{table}
\caption{Test \textbf{generalization on longer sequences}, concatenating $C$ consecutive sequences in the Hub5'00 corpora.}
\centering
\label{tab:concat-seqs}
\begin{adjustbox}{width=0.95\width,center}
\setlength{\tabcolsep}{0.3em}
\begin{tabular}{|S[table-format=2]|c|c||S[table-format=2.1]|S[table-format=2.1]|S[table-format=2.1]|}
  \hline
  {$C$} & \multicolumn{2}{c||}{Seq. length (secs)} & \multicolumn{3}{c|}{WER (\%)} \\
  \cline{4-6}
  & mean $\pm$ std & min-max & {Global} & \multicolumn{2}{c|}{Segmental} \\
  & & & {w/o sil.} & {w/o sil.} & {w/ sil.} \\
  \hline  \hline
  1 & $\phantom{0}2.91 \pm \phantom{0}2.57$ & $0.08-\phantom{0}15.51$ & 14.8 & 15.0 & 16.5 \\  \hline
  2 & $\phantom{0}8.10 \pm \phantom{0}5.19$ & $0.27-\phantom{0}63.97$ & 15.1 & 15.7 & 17.6 \\  \hline
  4 & $18.28 \pm \phantom{0}9.64$ & $0.27-100.46$ & 16.8 & 16.9 & 17.6 \\  \hline
  10 & $46.65 \pm 22.52$ & $0.33-198.10$ & 33.9 & 19.3 & 18.6 \\  \hline
  20 & $87.41 \pm 44.83$ & $0.89-282.78$ & 72.8 & 22.2 & 19.3 \\  \hline
\end{tabular}
\end{adjustbox}
\end{table}

We investigate performance on longer sequences than seen during training
by simply concatenating $C$ consecutive sequences during recognition.
We compare the global attention model to the segmental attention model
and collect results in \Cref{tab:concat-seqs}.
The segmental model generalizes much better than global attention model.
We also observe that the segmental model with explicit silence
generalizes slightly better than without explicit silence.
%
We believe that segmental models would generalize even better to very long sequences
if they would have seen some sentence concatenation in training.
Then, they should even improve the WER the longer the sequence becomes
due to more context,
as has been seen for language modeling
\cite{irie19:cross-sentence,tuske2020swbatt}.

\subsection{Overall comparison}

\begin{table}
\caption{\textbf{Overall comparison},
comparing global attention models, transducers and segmental models
from the literature
and our final best models,
different number of epochs (\#Ep),
seg.~model uses length model scales 1.0 or 0.7.}
\centering
\label{tab:final-results}
\begin{adjustbox}{width=0.72\width,center}
\setlength{\tabcolsep}{0.10em}
\begin{tabular}{|c|c|c|c|S[table-format=3]|c||>{\small}S[table-format=1.1]|>{\small}S[table-format=2.1]|S[table-format=2.1]||>{\small}S[table-format=2.1]|>{\small}S[table-format=2.1]|>{\small}S[table-format=2.1]|S[table-format=2.1]||>{\small}S[table-format=2.1]|>{\small}S[table-format=2.1]|S[table-format=2.1]|}
    \hline
    \multirow{3}{*}{Work} & \multirow{3}{*}{Model} & \multicolumn{2}{c|}{Label} & {\multirow{3}{*}{\#Ep}} & \multirow{3}{*}{LM} & \multicolumn{10}{c|}{WER (\%)}\\
    & & Type & \# & & & \multicolumn{3}{c||}{$\text{Hub}5^{00}$} & \multicolumn{4}{c||}{$\text{Hub}5^{01}$} & \multicolumn{3}{c|}{$\text{RT}^{03}$} \\
    & & & & & & {\scriptsize SWB} & {\scriptsize CH} & $\Sigma$ & {\scriptsize SWB} & {\scriptsize 2P3} & {\scriptsize 2P4} & $\Sigma$ & {\scriptsize SWB} & {\scriptsize FSH} & $\Sigma$\\
    \hline \hline
    \cite{zeineldeen2021ilm} & Gl.Att. & BPE & 534 & 33 & no &  &  & 14.6 &  &  & & 14.0 & & & 16.8 \\  \hline
    \multirow{2}{*}{\cite{zeyer2020:transducer}} & \multirow{2}{*}{Transd.} & \multirow{2}{*}{BPE} & \multirow{2}{*}{1k} & 25 & \multirow{2}{*}{no} & 9.4 & 18.7 & 14.1 &  &  & & 14.1 & & & 16.7\\   \cline{5-5} \cline{7-16}
    & & & & 50 & & 8.7 & 18.3 & 13.5 &  &  & & 13.3 & & & 15.6\\    \hline
    \cite{zhou:phoneme-transducer:2021} & Transd. & Phon. & 88 & 100 & yes &  &  & 11.2 &  &  & & 11.2 & & & \\     \hline
  \multirow{2}{*}{\cite{tuske21swb}} & \multirow{2}{*}{Gl.Att.} & \multirow{2}{*}{BPE} & \multirow{2}{*}{600} & {\multirow{2}{*}{$\phantom{0}70$}} & no & 6.7 & 13.0 & 9.9 & 7.1 & 9.2 & 13.7 & 10.1 & 15.7 & 9.1 & 12.5 \\  \cline{6-16}
    & & & & & yes & 5.5 & 11.2 & 8.4 & 6.1 & 7.7 & 11.4 & 8.5 & 12.6 & 7.0 & 9.9 \\
    \hline  \hline
    \multirow{4}{*}{Ours} & Gl.Att. & \multirow{4}{*}{BPE} & \multirow{4}{*}{1k} & {\multirow{3}{*}{$\phantom{0}25$}} & \multirow{4}{*}{no} & 9.1 & 20.4 & 14.8 & 10.4 & 13.1 & 18.6 & 14.2 & 20.4 & 12.9 & 16.8\\
    \cline{2-2} \cline{7-16}
    & Seg./1.0 & & & & & 9.3 & 19.0 & 14.2 & 10.3 & 13.1 & 18.1 & 14.0 & 19.9 & 12.8 & 16.5\\    \cline{2-2} \cline{7-16}
    & Seg./0.7 & & & & & 9.2 & 19.0 & 14.1 & 10.3 & 12.9 & 18.0 & 13.9 & 19.8 & 12.6 & 16.4 \\ \cline{2-2} \cline{5-5} \cline{7-16}
    & Seg./0.7 & & & 50 & & 8.8 & 18.5 & \best{13.7} & 10.0 & 12.4 & 17.4 & \best{13.4} & 19.4 & 12.4 & \best{16.0}\\  \hline
\end{tabular}
\end{adjustbox}
\end{table}

We collect our final results comparing
our best global-attention model to our best segmental-attention model
and other results from the literature
in \Cref{tab:final-results}.
Our segmental-attention model still uses an LSTM acoustic model,
no external language model yet,
and is not trained as long as other results
which explains the gap to some other work from the literature.
We see that segmental-attention overall performs
slightly better than global-attention.

\section{Conclusions}

We introduced a novel type of segmental-attention models
derived from the global-attention models.
Our investigation of the modeling of the segment boundaries
demonstrates and explains
why a good length model is important for such models:
the label model on its own still yields good scores
even for bad segment boundaries.
This can be explained due to the flexibility of attention
but similar observations has also been made for other segmental models
\cite{beck2018:segmental,beck2018:alignment}.
This is in contrast to the experience of HMMs
were the time-distortion penalties are less important.

In the end, our final segmental-attention model
improves over the global-attention model,
while at the same time satisfies all our motivations,
namely it is monotonic, allows for online-streaming,
potentially more efficient
(at least the neural model, the search is currently slower
but this is just for technical implementation reasons
and can be improved),
and it generalizes much better to long sequences.
We think the segmental model can still be improved
by not using $[{t_{s-1}+1},{t_s}]$ as the attention window
but instead some fixed-size window where $t_s$ is the center position.

When looking at the equivalence
of transducer and segmental models \cite{zhou2021segmentalvstransducer},
we note that the main difference of our segmental model
to common transducer models
is the use of attention on the segment,
while otherwise both models are conceptually similar.


\if\blind1%
\else%
\section{Acknowledgements}
\begingroup
\begin{small}
\begin{spacing}{0.8}
\setlength{\columnsep}{5pt}%
\setlength{\intextsep}{0pt}%
This work was partially supported by
NeuroSys which, as part of the initiative “Clusters4Future”,
is funded by the Federal Ministry of Education and Research BMBF (03ZU1106DA);
a Google Focused Award.
The work reflects only the authors' views and none of the funding parties is responsible for any use that may be made of the information it contains.
\end{spacing}
\end{small}
\endgroup
\fi%

\bibliographystyle{IEEEbib}

\let\OLDthebibliography\thebibliography
\renewcommand\thebibliography[1]{
  \OLDthebibliography{#1}
  \renewcommand{\baselinestretch}{0.1}\normalsize
  \setlength{\parskip}{0pt}
  \setlength{\itemsep}{2pt plus 0.07ex}
  \setstretch{0.8}  
}

\bibliography{strings,refs}

\end{document}